\documentclass[journal]{IEEEtran}

\newcommand{\etal}{\textit{et al}. }
\newcommand{\ie}{\textit{i}.\textit{e}.}

\usepackage[utf8]{inputenc} 
\usepackage[T1]{fontenc}    
\usepackage{hyperref}       
\usepackage{url}            
\usepackage{booktabs}       
\usepackage{amsfonts}       
\usepackage{nicefrac}       
\usepackage{microtype}      
\usepackage{amsmath}
\usepackage{bbm}
\usepackage{array}
\usepackage{wrapfig}
\usepackage{arydshln}
\usepackage[export]{adjustbox}
\usepackage{xparse}
\usepackage{multirow}
\usepackage{cite}
\usepackage[caption=false, font=footnotesize]{subfig}

\usepackage{tikz}
\let\svtikzpicture\tikzpicture
\def\tikzpicture{\noindent\svtikzpicture}
\usepackage{lipsum,adjustbox,cleveref}
\usetikzlibrary{calc,positioning}
\usetikzlibrary{shapes,arrows}

\usepackage{algpseudocode}
\usepackage{algorithm}
\algnewcommand\algorithmicforeach{\textbf{for}}
\algdef{S}[FOR]{ForEach}[1]{\algorithmicforeach\ #1\ \algorithmicdo}

\newcommand{\argmin}{\operatornamewithlimits{argmin}}

\NewDocumentCommand{\ShowInline}{v}{%
#1%
}

\makeatletter
\long\def\@makecaption#1#2{\ifx\@captype\@IEEEtablestring%
\footnotesize\begin{center}{\normalfont\footnotesize #1}\\
{\normalfont\footnotesize\scshape #2}\end{center}%
\@IEEEtablecaptionsepspace
\else
\@IEEEfigurecaptionsepspace
\setbox\@tempboxa\hbox{\normalfont\footnotesize {#1.}~~ #2}%
\ifdim \wd\@tempboxa >\hsize%
\setbox\@tempboxa\hbox{\normalfont\footnotesize {#1.}~~ }%
\parbox[t]{\hsize}{\normalfont\footnotesize \noindent\unhbox\@tempboxa#2}%
\else
\hbox to\hsize{\normalfont\footnotesize\hfil\box\@tempboxa\hfil}\fi\fi}
\makeatother
\title{Image Quality Assessment using Contrastive Learning}

\author{Pavan C. Madhusudana, Neil Birkbeck, Yilin Wang,  Balu Adsumilli and Alan C. Bovik 
	\thanks{P. C. Madhusudana and A. C. Bovik are with the Department of Electrical and
Computer Engineering, University of Texas at Austin, Austin, TX, USA (e-mail:
pavancm@utexas.edu; bovik@ece.utexas.edu). Neil Birkbeck, Yilin Wang
and Balu Adsumilli are with Google Inc. (e-mail: birkbeck@google.com; yilin@google.com; badsumilli@google.com).}}

\begin{document}

\maketitle

\begin{abstract}
We consider the problem of obtaining image quality representations in a self-supervised manner. We use prediction of distortion type and degree as an auxiliary task to learn features from an unlabeled image dataset containing a mixture of synthetic and realistic distortions. We then train a deep Convolutional Neural Network (CNN) using a contrastive pairwise objective to solve the auxiliary problem. We refer to the proposed training framework and resulting deep IQA model as the CONTRastive Image QUality Evaluator (CONTRIQUE). During evaluation, the CNN weights are frozen and a linear regressor maps the learned representations to quality scores in a No-Reference (NR) setting. We show through extensive experiments that CONTRIQUE achieves competitive performance when compared to state-of-the-art NR image quality models, even without any additional fine-tuning of the CNN backbone. The learned representations are highly robust and generalize well across images afflicted by either synthetic or authentic distortions. Our results suggest that powerful quality representations with perceptual relevance can be obtained without requiring large labeled subjective image quality datasets. The implementations used in this paper are available at \url{https://github.com/pavancm/CONTRIQUE}.
\end{abstract}


\section{Introduction}
\IEEEPARstart{I}{mage} Quality Assessment (IQA) pertains to the problem of quantifying and predicting human perceptual judgments of image quality. No-Reference (NR) or blind IQA is focused on estimating the quality of degraded images with no information about any pristine reference images or of the types of distortions that are present. The goal of NR-IQA models is to make robust and accurate quality predictions that correlate well with subjective judgments. The typical presence of multiple types of artifacts, as well as the influence of image content on perceived quality makes NR-IQA an interesting and challenging problem. NR-IQA has become a central technology for social media platforms such as Facebook, Instagram, Flickr etc. where millions of digital user-generated content (UGC) images are uploaded everyday. It is necessary to be able to objectively determine and control the quality of these digital photographs, and to guide subsequent processing tasks, such as compression \cite{yu2019predicting}.

NR-IQA has been a topic of intense interest among the research community for more than a decade, resulting in a variety of IQA datasets and objective models. Legacy IQA databases such as LIVE-IQA \cite{sheikh2006statistical}, CSIQ-IQA \cite{larson2010most} etc. have been influential in advancing the field of image quality prediction. These early datasets contain images with synthetic distortions, whereby a pristine high quality reference is artificially corrupted by commonly observed distortions such as blur, white noise, compression artifacts etc. However, a shortcoming of these datasets is that in most instances, a 'single' distortion type is applied on each image, whereas in reality images commonly are degraded by a combination of multiple distortions. To address this, various recent databases have been introduced that contain real, authentically distorted images \cite{ghadiyaram2015massive,hosu2020koniq,ying2019patches,fang2020perceptual}, typically captured by casual users with handheld camera devices. From the perspective of objective NR model design, it is desirable to obtain a model that can perform well on both synthetic and authentic distortions, so that it is applicable to any image regardless of the type of impairments it is afflicted with.

Well established NR-IQA models typically rely on parametric or learned approaches. Natural scene statistics (NSS) based models \cite{moorthy2011blind,saad2012blind,mittal2012no,mittal2013making} use features which are derived from statistical observations, and use them to predict visual quality. These kinds of algortithms have been very successful at analyzing synthetic artifacts, but their performance has proven to be limited when evaluated on images afflicted by unknown, often commingled authentic distortions. Over the last decade, the many successes of deep Convolutional Neural Networks (CNN) \cite{he2016deep,he2017mask,sun2018pwc} trained on large databases has motivated the development of many CNN based, data-driven IQA models have been proposed \cite{zhang2018blind,kim2016fully,zeng2017probabilistic,su2020blindly}. 

One barrier to the development of CNN based IQA models is the lack of availability of sufficiently large labeled IQA datasets. Annotating IQA datasets is an expensive and labor intensive process. Most available IQA datasets are too small to effectively train deep CNN models from scratch. Because of this, most CNN based IQA models utilize transfer learning, where the CNN is pretrained on a large dataset like ImageNet \cite{russakovsky2015imagenet}, then fine-tuned end-to-end on images with subjective quality judgments. Although fine-tuned models achieve impressive performances on both synthetic and authentic distortions, fine-tuning requires carefully chosen hyper-parameters that can vary with different IQA databases. Moreover, excessive fine-tuning can overfit the model on the training data limiting its generalizability.

Here we introduce a contrastive learning based IQA training framework aimed towards obtaining efficient image quality representations using \textit{unlabeled} datasets. Our ideas are motivated by the successes of unsupervised/self-supervised pretraining methods \cite{dosovitskiy2014discriminative,bojanowski2017,chen2020simple,he2020momentum} originally proposed for image classification problems.  We refer to the new model as \textbf{CONTR}astive \textbf{I}mage \textbf{QU}ality \textbf{E}valuator (CONTRIQUE). The salient characteristics of CONTRIQUE are as follows:
\begin{enumerate}
    \item We use prediction of distortion type and degree as an auxiliary task to train a deep CNN from scratch. Training is done on an unlabeled dataset containing both synthetic and authentic distortions, using a contrastive objective function.
    \item To learn robust representations, multiscale and quality preserving transformations are performed on the unlabeled data during training.
    \item During testing, the weights of the deep CNN are frozen, and features from this network are mapped to quality scores using a simple linear regressor. Quality predictions produced by CONTRIQUE are shown to be competitive with those of state-of-the-art (SOTA) IQA models across multiple databases. This is accomplished with no additional fine-tuning of the CNN backbone.
    \item The CONTRIQUE training framework is simple, and results in in highly generalizable representations that perform well on both synthetic and realistic distortions. Additionally, we show that the CONTRIQUE features can be easily extended to the Full-Reference (FR) IQA problem with no additional training of the CNN backbone.
\end{enumerate}

The rest of the paper is organized as follows: In Section \ref{sec:prior_work} we discuss prior methods related to IQA and self-supervised learning. In Section \ref{sec:Method} we provide a detailed description of the design of CONTRIQUE. Section \ref{sec:experiments} analyzes and compares various experimental results of CONTRIQUE, and we conclude in Section \ref{sec:conclusion}.

\section{Related Work}
\label{sec:prior_work}
In this section we review related work from the literature concerning NR-IQA and self-supervised learning.

\subsection{NR-IQA Models}
Blind image quality prediction is a challenging problem due to the diverse types of artifacts involved. The influence of image content on different distortion types adds additional complexity to the problem. Over the past decade, considerable research effort has been expended on designing NR-IQA models, with the goal of obtaining quality predictions that have high correlations against human judgements. NR models can be broadly categorized based on the design methodology - traditional/hand-crafted models, and deep CNN based models. Most prior models pursue a design philosophy of having a feature extraction framework followed by a regressor to map features to quality values. In traditional models, feature extraction is accomplished by modeling the image artifacts, Natural Scene Statistics (NSS) based models are a popular example employing this approach. NSS models extract features from a transform domain, where deviations from expected statistical regularities due to distortions are predictive of quality. NSS models include DIIVINE \cite{moorthy2011blind}, which employs steerable pyramids, BLIINDS \cite{saad2012blind}, which uses DCT coefficients, and BRISQUE \cite{mittal2012no} and NIQE \cite{mittal2013making}, which use mean subtracted contrast normalized coefficients (MSCN) to obtain quality aware features. In CORNIA \cite{ye2012unsupervised} and HOSA \cite{xu2016blind}, a visual codebook constructed from local patches is used to obtain quality representative features. Although traditional models achieve impressive performances when evaluated on images with synthetic distortions, their capabilities are often limited when tested on images containing realistic distortions and combinations of them.

The successes of deep learning on many computer vision tasks \cite{he2016deep,he2017mask,sun2018pwc} has inspired a large number of CNN-based NR-IQA models. The motivation behind using CNN is to obtain reliable semantic features from deep architectures, then perform appropriate modifications to adapt them for quality prediction. Due to a lack of large scale data pertaining to image quality, the majority of CNN-based models use transfer learning techniques, whereby a pretrained model (usually pretrained on ImageNet \cite{russakovsky2015imagenet}) is fine-tuned using ground-truth image quality labels. In \cite{kim2017deep}, it was shown that features obtained from pretrained CNN architectures like Resnet \cite{he2016deep} can be particularly effective in capturing authentic distortions. In \cite{zhang2018blind}, two separate CNNs are employed to account for synthetic and authentic artifacts, respectively. Kim \etal \cite{kim2016fully} employed FR-IQA maps as intermediate regression targets during training. Zeng \etal \cite{zeng2017probabilistic} used a statistical distribution of subjective scores when training which led to faster convergence and resulted in superior quality estimates. Su \etal \cite{su2020blindly} proposed an adaptive hyper network architecture to separate quality prediction from content understanding. Ying \etal \cite{ying2019patches} demonstrated that training with both image and patch quality scores can significantly boost model performance. The PaQ-2-PiQ algorithm developed by these authors also benefited by the availability of an unusually large subjective database of realistically distorted images. All these models rely on specific supervised fine-tuning mechanisms in order to achieve improved performance. In contrast, our work focuses on \textit{unsupervised} feature learning with no fine-tuning procedures.

\subsection{Self-Supervised Learning}
Self-supervised learning or unsupervised pretraining aims at obtaining representations using unlabeled data. These techniques derive useful representations by exploiting existing structural information available in the image data. Recent SOTA methods rely on instance discrimination task, in which each image and augmented versions of it are treated as a single class \cite{dosovitskiy2014discriminative,chen2020simple,he2020momentum}. Another form of self-supervision involves learning features through auxiliary tasks (different but related to the original task) for which data is abundant, and which requires no annotations. Examples of these self-supervised tasks include rotation prediction \cite{gidaris2018unsupervised}, obtaining color images from grayscale and vice versa \cite{zhang2016colorful,larsson2017colorization}, and inpainting \cite{pathak2016context}. Liu \etal \cite{liu2019exploiting} proposed an NR-IQA model using image ranking as an auxiliary task, and achieved competitive performance on datasets with synthetic artifacts. Here we use discrimination of distortion types and degrees, which is related to quality assessment, as a self-supervision task and then we use the learned representations for image quality prediction.

\section{Method}
\label{sec:Method}

\tikzstyle{block} = [draw, fill=blue!20, rounded corners, text centered,  minimum height=3em, minimum width=4em]
\tikzstyle{block2} = [draw, rounded corners, text centered,  minimum height=3em, minimum width=4em]
\tikzstyle{circle1} = [draw, fill=blue!20, circle, text centered,  minimum size=0.5em]
\tikzstyle{input} = [coordinate]
\tikzstyle{output} = [coordinate]
\tikzstyle{pinstyle} = [pin edge={to-,thin,black}]
\pgfdeclarelayer{background}
\pgfdeclarelayer{foreground}
\pgfsetlayers{background,main,foreground}

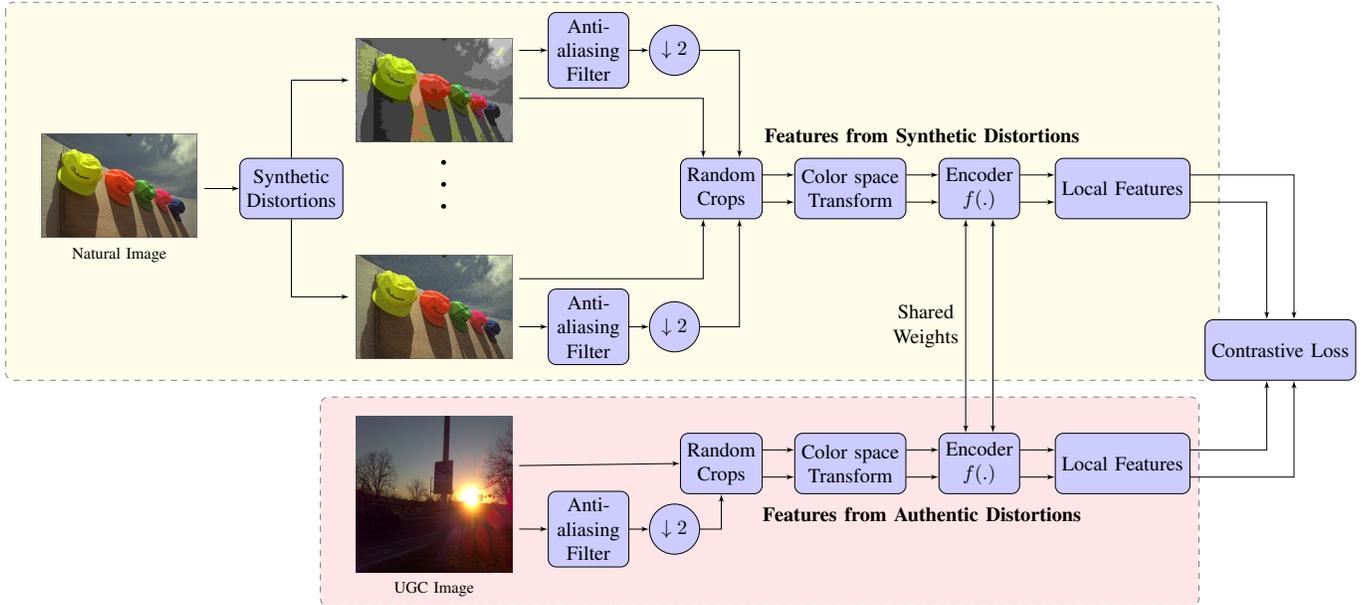
\begin {figure*}[t]
\captionsetup[subfigure]{justification = centering,labelformat=empty,position=bottom}
\begin{minipage}[b]{1\linewidth}
	\centering
	\resizebox{1\textwidth}{!}{%
		\begin{tikzpicture}[auto, node distance=2cm,>=latex']
		
		\node (nat_im) [left = 1.5cm,align=center]{\subfloat[Natural Image]{\includegraphics[width = 0.15\linewidth]{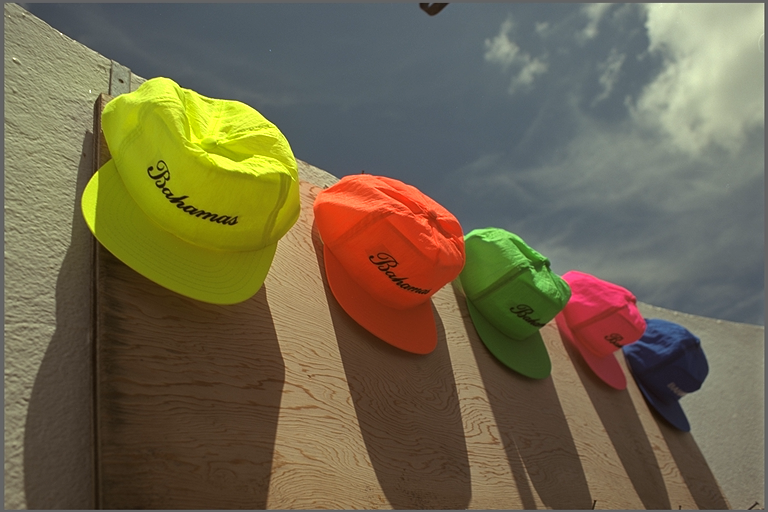}}};
		
		\node (syn_dist) [block, right of=nat_im,node distance=3cm, align=center]{Synthetic \\ Distortions};
		
		\node (syn1) [above right = 0.15cm and 0.1cm of syn_dist, align=center]{\subfloat{\includegraphics[width=0.15\linewidth]{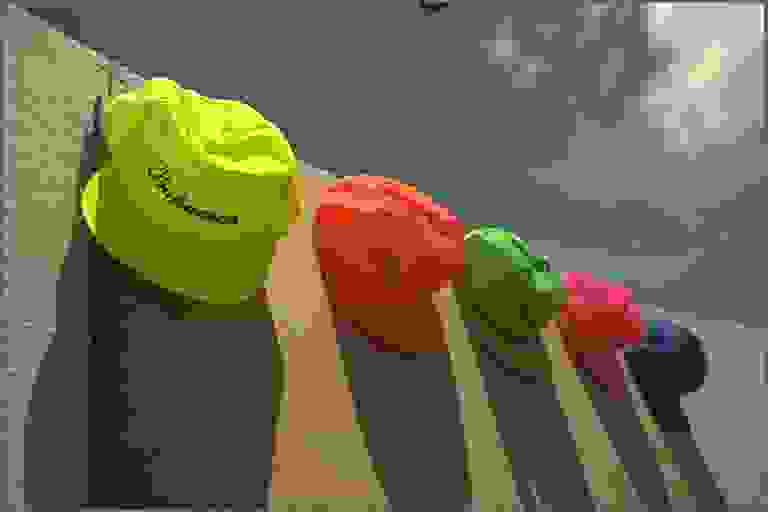}}};
		
		\node (syn2) [below right = 0.15cm and 0.1cm of syn_dist, align=center]{\subfloat{\includegraphics[width=0.15\linewidth]{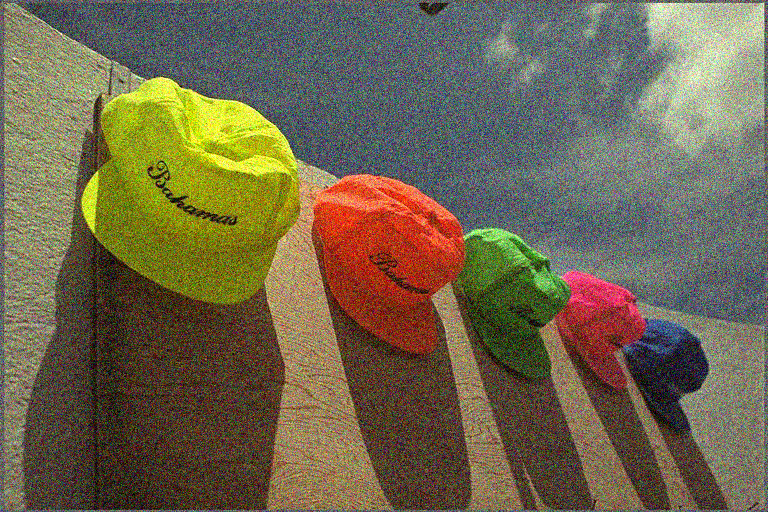}}};
		
		\node (r_patch1) [block, right of=syn_dist, node distance = 7.5cm, align=center] {Random \\ Crops};
		\node (alias1)[block, below right = -1.35cm and 0.5cm of syn2, align=center] {Anti-\\aliasing\\Filter};
		\node (alias12)[block, above right = -1.35cm and 0.5cm of syn1, align=center] {Anti-\\aliasing\\Filter};
		\node (ds1) [circle1, right of=alias1, node distance = 1.5cm, align=center] {$\downarrow 2$};
		\node (ds12) [circle1, right of=alias12, node distance = 1.5cm, align=center] {$\downarrow 2$};
		\node (ct1) [block, right of=r_patch1, node distance = 2.25cm, align=center] {Color space \\ Transform};
		\node (enc1) [block, right of=ct1, node distance = 2.25cm, align=center] {Encoder \\ $f(.)$};
		\node (local1) [block, right of=enc1, node distance = 2.5cm, align=center] {Local Features};
		
		\node (auth_im) [below of=syn2, node distance = 3.5cm, align=center] {\subfloat[UGC Image]{\includegraphics[width = 0.15\linewidth]{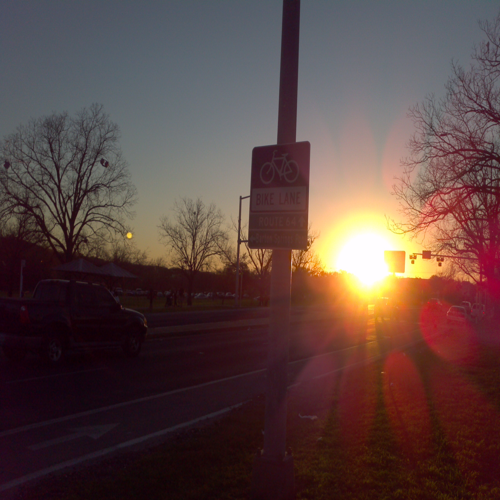}}};
		\node (r_patch2) [block, below of=r_patch1, node distance = 4.8cm, align=center] {Random \\ Crops};
		\node (alias2)[block, below right = -2cm and 0.5cm of auth_im, align=center] {Anti-\\aliasing\\Filter};
		\node (ds2) [circle1, right of=alias2, node distance = 1.5cm, align=center] {$\downarrow 2$};
		
		\node (ct2) [block, right of=r_patch2, node distance = 2.25cm, align=center] {Color space \\ Transform};
		\node (enc2) [block, right of=ct2, node distance = 2.25cm, align=center] {Encoder \\ $f(.)$};
		\node (local2) [block, right of=enc2, node distance = 2.5cm, align=center] {Local Features};
		
		\node (c_loss) [block, below right = 1.75cm and 0.25cm of local1, align=center] {Contrastive Loss};
		
		\path (syn1.south) -- (syn2.north) node[font=\Huge, midway, sloped]{$\dots$};
		\draw [->] (nat_im) -- node{}(syn_dist);
		\draw [->] (syn_dist) |- node{}(syn1);
		\draw [->] (syn_dist) |- node{}(syn2);
		\draw [->] ([yshift=2ex]syn2.east) -| node{}([xshift=-2ex]r_patch1.south);
		\draw [->] ([yshift=-3.3ex]syn2.east) -- node{}(alias1.west);
		\draw [->] (alias1.east) -- node{}(ds1.west);
		\draw [->] ([yshift=-2ex]syn1.east) -| node{}([xshift=-2ex]r_patch1.north);
		\draw [->] ([yshift=3.3ex]syn1.east) -- node{}(alias12.west);
		\draw [->] (alias12.east) -- node{}(ds12.west);
		\draw [->] ([yshift=3.5ex]auth_im.east) -- node{}(r_patch2.west);
		\draw [->] ([yshift=-3.5ex]auth_im.east) -- node{}(alias2.west);
		\draw [->] (alias2.east) -- node{}(ds2.west);
		
		\draw [->] (ds1) -| node{}([xshift=2ex]r_patch1.south);
		\draw [->] (ds12) -| node{}([xshift=2ex]r_patch1.north);
		\draw [->] (ds2) -| node{}(r_patch2.south);
		\draw [->] ([yshift=-1.5ex]r_patch1.east) -- node{}([yshift=-1.5ex]ct1.west);
		\draw [->] ([yshift=1.5ex]r_patch1.east) -- node{}([yshift=1.5ex]ct1.west);
		\draw [->] ([yshift=-1.5ex]r_patch2.east) -- node{}([yshift=-1.5ex]ct2.west);
		\draw [->] ([yshift=1.5ex]r_patch2.east) -- node{}([yshift=1.5ex]ct2.west);
		\draw [->] ([yshift=-1.5ex]ct1.east) -- node{}([yshift=-1.5ex]enc1.west);
		\draw [->] ([yshift=1.5ex]ct1.east) -- node{}([yshift=1.5ex]enc1.west);
		\draw [->] ([yshift=-1.5ex]ct2.east) -- node{}([yshift=-1.5ex]enc2.west);
		\draw [->] ([yshift=1.5ex]ct2.east) -- node{}([yshift=1.5ex]enc2.west);
		\draw [->] ([yshift=-1.5ex]enc1.east) |- node{}([yshift=-1.5ex]local1.west);
		\draw [->] ([yshift=1.5ex]enc1.east) |- node{}([yshift=1.5ex]local1.west);
		\draw [->] ([yshift=-1.5ex]enc2.east) |- node{}([yshift=-1.5ex]local2.west);
		\draw [->] ([yshift=1.5ex]enc2.east) |- node{}([yshift=1.5ex]local2.west);
		\draw [->] ([yshift=-1.5ex]local1.east) -| node{}([xshift=-1.5ex]c_loss.north);
		\draw [->] ([yshift=1.5ex]local1.east) -| node{}([xshift=1.5ex]c_loss.north);
		\draw [->] ([yshift=-1.5ex]local2.east) -| node{}([xshift=1.5ex]c_loss.south);
		\draw [->] ([yshift=1.5ex]local2.east) -| node{}([xshift=-1.5ex]c_loss.south);
		
		\draw[<->] ([xshift=-1.5ex]enc1.south) -- node[left, align=center]{Shared \\ Weights}([xshift=-1.5ex]enc2.north);
		\draw[<->] ([xshift=1.5ex]enc1.south) -- node{}([xshift=1.5ex]enc2.north);
		
		\begin{pgfonlayer}{background}
		\path (nat_im.west)+(-0.5,3.25) node (a) {};
		\path (local1.east |- syn2.south)+(+0.5,-0.25) node (c) {};
		
		\path[fill=yellow!10,rounded corners, draw=black!50, dashed](a) rectangle (c);           
		
		\end{pgfonlayer}
		
		\begin{pgfonlayer}{background}
		\path (auth_im.west)-|+(-0.5,1.75) node (d) {};
		\path (local2.east |- ds2.south)+(+0.15,-0.9) node (e) {};
		
		\path[fill=red!10,rounded corners, draw=black!50, dashed](d) rectangle (e);           
		
		\end{pgfonlayer}
		
		\path (local1.north) +(-3.5,0.35) 
		node (asrs) {\textbf{Features from Synthetic Distortions}};
		
		\path (local2.south) +(-3.5,-0.35) 
		node (asrs) {\textbf{Features from Authentic Distortions}};
		\end{tikzpicture}
	}
\end{minipage}
\centering\caption{Illustration of training pipeline of the CONTRIQUE Framework}
\label{fig:overview}
\end{figure*}

Our method is a transform domain approach where a transformation $f:\mathbb{R}^{3 \times H \times W} \mapsto \mathbb{R}^d$ maps an image $x$ to a representation $h$. Bandpass transformations such as wavelet-like decompositions are often used to model the responses of visual neurons in primary visual cortex that are tuned to visual stimuli having specific spatial locations, frequencies, and orientations. Traditional NR-IQA models have been based on band-pass transformations such as the DCT \cite{saad2012blind}, steerable pyramids \cite{moorthy2011blind}, local mean-subtraction \cite{mittal2012no,mittal2013making} and so on, have been highly effective at predicting perceptual quality. Recently, transformations induced by deep CNNs have demonstrated remarkable efficiency at capturing perceptual image artifacts \cite{zhang2018blind,su2020blindly,zeng2017probabilistic}.

Here, our goal is to learn robust representations that can be used to predict image quality, without employing any ground-truth quality scores during training. Our proposed training pipeline is illustrated in Fig. \ref{fig:overview}. In the following sections each module present in the framework is discussed in detail. 

\subsection{Auxiliary Task}
An auxiliary task to learning problem is an alternate but closely related task, for which the ground-truth labels are known or can easily be obtained. In this approach, model is trained to solve an auxiliary problem, then during the inference stage, the trained model is evaluated on the original task. In the case of IQA, the goal is to obtain discriminative representations that can distinguish different types of distortions, as well as the degrees of degradations. Thus, we transform the IQA representation learning problem to a classification problem, where each class consists of images having a similar type of distortion, as well as similar degree of quality degradation. The goal of the auxiliary task is to learn features that can differentiate images into distortion dependent classes, similar to \cite{zhang2018blind,kim2020dynamic}, which employ a cross-entropy objective during training to achieve this.

Let a pristine high quality image $x$ be degraded by a distortion $d^i, i \in \{1,\ldots ,D\}$ with degradation degree $l^{ij}, j \in \{1,\ldots,L^i\}$ resulting in a distorted image $\Tilde{x}^{j}_i$. Here, $D$ and $L^i$ correspond to the number of distortion types and degradation degrees, respectively. For a given $\Tilde{x}^{j}_i$, the task of the model is to identify $d^i$ and $l^{ij}$. This task translates to a classification problem having $\sum_{i=1} ^D L^i + 1$ classes (total number of degradation levels + one pristine image). Motivated by the successes of using contrastive loss \cite{chen2020simple,he2020momentum} for learning representations, we incorporate a similar technique into the CONTRIQUE framework. To extract embeddings, we define a deep model consisting of two parts : an encoder and projector. The encoder can be any popular CNN architecture such as VGG \cite{simonyan2014very}, Resnet \cite{he2016deep} etc., with any fully connected terminal layer removed. The projector is a multi-layer perceptron (MLP) that reduces the dimensionality of the representation produced by the encoder. Let $f(.)$ and $g(.)$ denote the deep encoder network and the projector network respectively. For a given image $x \in \mathbb{R}^{3 \times H\times W}$
\begin{align}
    h = f(x), \qquad z = g(h) = g(f(x)) \quad h \in \mathbb{R}^D, z \in \mathbb{R}^K
\end{align}
where $h$ is the $D$-dimensional output from the encoder. Similar to \cite{chen2020simple,he2020momentum} the encoder output $h$ is $L_2$ normalized before being fed to the projector network. Note that the output of the entire model $z$ is a $K$-dimensional vector (where $K$ is a hyperparameter in this design). The goal is to obtain similar representations $z$ of images belonging to the same class. The similarity between a pair of representations is measured using the dot product $\phi(a,b) = a^T b/||a||_2 ||b||_2$. The loss function is a normalized temperature-scaled cross entropy (NT-Xent), and for image $x_i$ is defined as 
\begin{align}
    \mathcal{L}_i ^{syn} = \frac{1}{|P(i)|} \sum_{j \in P(i)} -\log \frac{\exp(\phi(z_i,z_j)/\tau)}{\sum_{k=1}^N \mathbbm{1}_{k \neq i}\exp(\phi(z_i,z_k)/\tau)},
    \label{eqn:cont_loss_sup}
\end{align}
where $N$ is the number of images present in the batch, $\mathbbm{1}$ is the indicator function, $\tau$ is the temperature parameter, $P(i)$ is a set containing image indices belonging to the same class as $x_i$ (but excluding the index $i$) and $|P(i)|$ is its cardinality. The objective function (\ref{eqn:cont_loss_sup}) is similar to the supervised contrastive loss proposed in \cite{khosla2020supervised}. However, in \cite{khosla2020supervised} it was employed in the context of image classification with ground-truth labels, while in our design we incorporate prior knowledge of synthetic distortions as class labels. Another observation that can be made about the objective described in (\ref{eqn:cont_loss_sup}) is that it measures pairwise similarities between every pair of images in a batch. This pairwise loss computation is a key characteristic that differentiates it from the traditional cross-entropy loss.

\subsection{Multiscale Learning and Cropping}
Images are inherently multi-scale, as are distortions of them, and perceived image quality is influenced by both local characteristics as well as global details. Prior IQA models \cite{wang2003multiscale,moorthy2011blind,mittal2012no,mittal2013making} have attempted to simulate the functionality of front-end visual processing in the brain by employing multi-scale representations when predicting quality. CNN based IQA models \cite{zhang2018unreasonable,su2020blindly}, which use multi-scale features, are able to achieve remarkable efficiency in capturing visual quality. In CONTRIQUE, we employ two scales : native/full resolution, and half-scale resolution obtained by downsampling by a factor of two along both dimensions. To avoid aliasing artifacts, an anti-aliasing filter is used before downsampling as shown in Fig. \ref{fig:overview}. Note that the aspect ratio is preserved in this resizing operation, since modifying this ratio can affect the quality of the underlying image.

The images are then subjected to random cropping where the input images are cropped to a random fixed size $M \times M$. A simplifying assumption we make here is that the cropped version inherits the same distortion class as the original version. Although the cropped version need not represent the same perceived quality as the original image, we presume that the distortion class remains nearly the same and is unaffected by the cropping operation. For each input image, two random crops are obtained, one each at full-scale and half-scale. For cases where the size of the image was smaller than $M \times M$, the entire image was employed with zero padding to maintain the same resolution. Additionally, cropping provides images of fixed resolution in a batch, which is essential when training deep networks, since training with variable resolutions can be challenging and unstable \cite{ying2019patches}.

\subsection{Quality Preserving Transformations/Augmentations}
\label{sec:transform}
The goal of the objective function in (\ref{eqn:cont_loss_sup}) is to learn image embeddings that demonstrate discriminative behavior among images belonging to different classes, and at the same time exhibit invariance to quality preserving transformations. Image operations that do not modify image quality we collectively refer to as quality preserving transforms. In the CONTRIQUE framework, we employ two transforms: horizontal flipping and color space conversion. 

The motivation behind using different color spaces is to extract complementary quality information that can be present across different domains. In our proposed framework, we employ 4 color spaces: RGB, LAB, HSV and grayscale. Each of these color spaces have different types of perceptual relevance and have earlier been used in NSS based models \cite{ghadiyaram2017perceptual,tu2021rapique} to obtain quality features. We also employ a band-pass transform, obtained using local Mean-Subtraction (MS). MS coefficients have been shown to capture statistical deviations arising due to distortions in images \cite{bampis2017speed, bampis2018spatiotemporal,madhusudana2021st}. In the training pipeline shown in Fig. \ref{fig:overview}, the color space is randomly chosen for each crop of the input image. By employing different color spaces during training, as we show in Sec. \ref{sec:results_colorspace} that using any color space during testing results in similar representations, making CONTRIQUE invariant to color spaces. Note that we avoid employing aggressive augmentation techniques such as color jitter, Gaussian blur, random-resize, MixUp \cite{zhang2017mixup}, AutoAugment \cite{cubuk2019autoaugment} etc. as these methods modify distortion information and hence are not quality preserving.

\subsection{Realistic Distortions}
Prior knowledge about synthetic distortions was employed in the contrastive objective (\ref{eqn:cont_loss_sup}) to learn image quality embeddings. However, for images containing realistic distortions, such as User Generated Content (UGC) images, information regarding the distortion types is usually not available. Being able to handle authentic distortions is quite important since several hundred billion images are uploaded and shared to social media sites like Facebook, Instagram, YouTube etc. every year. UGC images, which are often afflicted by diverse mixtures of unknown distortions. Thus, the synthetic distortion classes assumed in (\ref{eqn:cont_loss_sup}) are not applicable to UGC images. In the CONTRIQUE framework, each UGC image is treated as a unique class obtained by a distinctive combination of multiple distortions, separate and distinct from other UGC images, as well as from images with synthetic artifacts. Thus, for a given UGC image $x_i$, only its scaled (and transformed) version $x_j$ belongs to the same class. To reflect this modification, we redefine the contrastive objective as 
\begin{align}
    \mathcal{L}_i ^{UGC} =  -\log \frac{\exp(\phi(z_i,z_j)/\tau)}{\sum_{k=1}^N \mathbbm{1}_{k \neq i}\exp(\phi(z_i,z_k)/\tau)}.
    \label{eqn:cont_loss_ugc}
\end{align}
This objective is similar to the one used in \cite{chen2020simple,he2020momentum} for the instance discrimination task. As detailed in Sec. \ref{sec:transform}, for each image there exists two transformed versions, at full-scale and half-scale. Thus, there are at least two datasamples belonging to the same class making the objective (\ref{eqn:cont_loss_ugc}) non-zero. The expression described in (\ref{eqn:cont_loss_ugc}) can also be considered as the special case of (\ref{eqn:cont_loss_sup}) where $P(i) = \{j\}$, \ie{} in a given batch only image $x_j$ belongs to the same class as $x_i$. The overall training objective is then
\begin{align}
    \mathcal{L} = \frac{1}{N} \sum_{i = 1} ^N \mathbbm{1}_{(x_i \notin UGC)} \mathcal{L}_i ^{syn} + \mathbbm{1}_{(x_i \in UGC)}\mathcal{L}_i ^{UGC},
    \label{eqn:cont_loss_total}
\end{align}
where $N$ is the number of images present in the batch, and $\mathbbm{1}$ is the indicator function determining whether the input image is non-synthetically distorted (UGC). During training, to avoid bias, we randomly sampled equal numbers of synthetic and authentically distorted images to form each batch, at each iteration.

\subsection{Patch Features}
Local details present in image patches play a significant role in determining global picture quality. Several patchwise learning based models have been proposed in the literature \cite{kang2014convolutional,kim2017deep} and shown to be effective for quality prediction. In order to capture distortion and image quality characteristics in a more granular fashion, we partitioned each input image into non-overlapping patches of size $P \times P$. These patches were then fed to the encoder module to obtain local features, and these representations are used in the contrastive objective function (\ref{eqn:cont_loss_total}). Similar to cropping operation, we assume the patches inherit the distortion class labels from the original image for both the synthetic as well as the realistically distorted images. Note that patches need not inherit the perceived quality of the original version, only the distortion class is presumed to be same. In addition to capturing local spatial neighborhood information, including patches provides increased number of data samples for every class, which can be beneficial for gradient descent based learning schemes. 

\begin{table*}[t]
\caption{Performance comparison of CONTRIQUE against different NR models on IQA databases containing \textbf{authentic} distortions. Models are categorized based on the type of feature extraction used. In each column, the first and second best models are boldfaced. Entries marked '-' denote that the results are not available.}
\label{table:authentic_IQA}
\centering
    \begin{tabular}{|c||c||cc|cc|cc|cc|}
        \hline
        \multirow{2}{*}{Method} & \multirow{2}{*}{Model Type} & \multicolumn{2}{|c|}{KonIQ \cite{hosu2020koniq}}  & \multicolumn{2}{|c|}{CLIVE \cite{ghadiyaram2015massive}} & \multicolumn{2}{|c|}{FLIVE \cite{ying2019patches}} & \multicolumn{2}{|c|}{SPAQ \cite{fang2020perceptual}}\\ \cline{3-10}
        ~ & ~ & SROCC$\uparrow$ & PLCC$\uparrow$ & SROCC$\uparrow$ & PLCC$\uparrow$ & SROCC$\uparrow$ & PLCC$\uparrow$ & SROCC$\uparrow$ & PLCC$\uparrow$ \\ \hline \hline
        BRISQUE \cite{mittal2012no} & \multirow{2}{*}{\shortstack[c]{Traditional/ Handcrafted \\ Features}} & 0.665 & 0.681 & 0.608 & 0.629 & 0.288 & 0.373 & 0.809 & 0.817\\
        NIQE \cite{mittal2012no} & ~ & 0.531 & 0.538 & 0.455 & 0.483 & 0.211 & 0.288 & 0.700 & 0.709 \\ \hline
        CORNIA \cite{ye2012unsupervised} & \multirow{2}{*}{\shortstack[c]{Codebook-based \\ Features}} & 0.780 & 0.795 & 0.629 & 0.671 & - & - & 0.709 & 0.725 \\
        HOSA \cite{xu2016blind} & ~ & 0.805 & 0.813 & 0.640 & 0.678 & - & - & 0.846 & 0.852 \\ \hline
        DB-CNN \cite{zhang2018blind} & \multirow{4}{*}{\shortstack[c]{Supervised pretraining and \\ supervised fine-tuning}} & 0.875 & 0.884 & 0.851 & 0.869 & 0.554 & \textbf{0.652} & 0.911 & 0.915\\
        PQR \cite{zeng2017probabilistic} & ~ & 0.880 & 0.884 & \textbf{0.857} & \textbf{0.882} & - & - & - & -\\
        PaQ-2-PiQ \cite{ying2019patches} & ~ & 0.870 & 0.880 & 0.840 & 0.850 & 0.571 & 0.623 & - & -\\ 
        HyperIQA \cite{su2020blindly} & ~ & \textbf{0.906} & \textbf{0.917} & \textbf{0.859} & \textbf{0.882} & 0.535 & 0.623 & \textbf{0.916} & \textbf{0.919} \\ \hline
        \multirow{2}{*}{Resnet-50 \cite{he2016deep}} & \multirow{2}{*}{\shortstack[c]{Supervised pretraining and \\ Linear Regression}} & \multirow{2}{*}{0.888} & \multirow{2}{*}{0.904} & \multirow{2}{*}{0.781} & \multirow{2}{*}{0.809} & \multirow{2}{*}{\textbf{0.595}} & \multirow{2}{*}{\textbf{0.648}} & \multirow{2}{*}{0.904} & \multirow{2}{*}{0.909}\\
        ~ & ~ & ~ & ~ & ~ & ~ & ~ & ~\\ \hline
        \multirow{2}{*}{CONTRIQUE} & \multirow{2}{*}{\shortstack[c]{Unsupervised pretraining and \\ Linear Regression}} & \multirow{2}{*}{\textbf{0.894}} & \multirow{2}{*}{\textbf{0.906}} & \multirow{2}{*}{0.845} & \multirow{2}{*}{0.857} &  \multirow{2}{*}{\textbf{0.580}} & \multirow{2}{*}{0.641} & \multirow{2}{*}{\textbf{0.914}} & \multirow{2}{*}{\textbf{0.919}}\\
        ~ & ~ & ~ & ~ & ~ & ~ & ~ & ~ & ~ & ~\\
        \hline
    \end{tabular}
\end{table*}


\subsection{Evaluating Representations}
We evaluate the learned representations by applying them to the quality prediction problem, using the correlations of human judgements against predicted quality scores as a proxy for representation quality. Once the training is complete, the projector network $g(.)$ is discarded and the outputs of encoder network $h = f(x)$ are used as image representations. We use a regularized linear regressor (ridge regression) trained on top of the frozen encoder network. This is similar to the linear evaluation protocol used in \cite{zhang2016colorful,oord2018representation,bachman2019learning} to evaluate the classification accuracy of self-supervised models. The regression weights are learned on a suitable IQA database containing ground-truth quality scores. The expression for ridge regression is given by
\begin{align}
    y = W h, \quad W^* = \argmin_{W} \sum_{i=1} ^N (GT_i - y_i)^2 + \lambda \sum_{j=1} ^M W_j^2,
    \label{eqn:regressor}
\end{align}
where $GT$ denotes ground-truth quality scores, $y$ predicted scores, $W$ is a trainable vector having same dimensions as $h$, $\lambda$ is the regularization parameter, $M$ is number of dimensions of $h$, and $N$ is the number of images present in the training set. Similar to training, we follow multiscale convention, and features are computed at two resolutions : full-scale and half-scale, and the final representation is a concatenation of both scales. During evaluation, all the representations are calculated at the native resolution of the input image, and no additional data augmentations are performed. Note that we do not perform any additional \textit{fine-tuning} whereby encoder weights would have been modified using the supervision of ground-truth quality scores. Although fine-tuning can potentially yield better performance, we avoid it as it alters the learned encoder weights, and it would not be a true indicator of the efficiency of the unsupervised training process. Additionally, we show in Sec. \ref{sec:correlation} that even without fine-tuning, CONTRIQUE achieves competitive performance as compared with SOTA IQA models.

\section{Experiments and Results}
\label{sec:experiments}
In this section we evaluate the performance of CONTRIQUE by conducting a series of experiments. We will first describe the experimental settings, evaluation protocol and compared methods. Then we explain how we evaluated CONTRIQUE against SOTA IQA models on multiple IQA databases. We perform a variety of ablation experiments to analyze the significance of distortion types present in the training data, as well as the importance of using different color spaces during training. Additionally, we study the generalizability of the CONTRIQUE features by performing cross-dataset testing.

\subsection{Experimental Settings}
\subsubsection*{\textbf{Training Data}}
The training data contains a combination of images impaired by synthetic and authentic distortions.
\begin{itemize}
    \item Synthetic Distortions : We utilized the KADIS dataset \cite{lin2020deepfl} to learn synthetic artifacts. The KADIS dataset contains 700k distorted images obtained from 140k pristine images and contains no subjective quality scores. There are 25 different types of distortions with each distortion spanning 5 degrees of degradation. The distortion types include compression, white noise, blur etc. Interested readers can refer to \cite{lin2020deepfl} for more details about the distortions present in this dataset. As there are $D=25$ distortions and $L^i=5$ degrees for each distortion type, a total of $25 \times 5 + 1$ (pristine image) $ = 126$ synthetic classes are used in the contrastive objective (\ref{eqn:cont_loss_total}).
    \item Authentic Distortions : We use a combination of 4 datasets aimed at capturing realistic distortions. (a) The AVA dataset \cite{murray2012ava} contains 255k images originally designed for aesthetic visual analysis. (b) The COCO dataset \cite{lin2014microsoft} contains 330k images designed to assist learning the detection and segmentation of objects occurring in common contexts. (c) The CERTH-Blur dataset \cite{mavridaki2014no} contains 2450 images captured with realistic blur. (d) The VOC \cite{everingham2010pascal} contains 33k images initially proposed for object recognition task. We discarded all the labels (if any) present in these datasets before training. 
\end{itemize}
Thus, a total of 1.3 million images were used to train CONTRIQUE.\\

\subsubsection*{\textbf{Training Details}}
We used a Resnet-50 \cite{he2016deep} architecture as the encoder network $f(.)$ and included 2 layers of MLP as the projector network $g(.)$. The hidden layers of MLP contained 2048 neurons each. The dimension of the final output $z$ was chosen to be $K = 128$. The CONTRIQUE framework is fairly generic in nature, and can easily be extended to other CNN based architectures. The training was done using a batch size of $N = 512$, with $256$ images randomly chosen from the synthetic distortion set and the rest authentically distorted. The sampled images were cropped to square blocks of size $M = 256$. These crops constitute approximately 50\% of the original dimensions of the images present in the training data. When extracting patch features, patches of size $P = 64$ were used, resulting in 4 patches from each input image. Patch features were computed by using an adaptive average pooling layer at the end of the encoder. The temperature parameter used in (\ref{eqn:cont_loss_sup}) and (\ref{eqn:cont_loss_ugc}) was fixed at $\tau = 0.1$. The model was trained from scratch for 25 epochs using a stochastic gradient descent (SGD) optimizer with initial learning rate of $0.6$. Furthermore, the learning rate was subjected to a linear warmup for the first two epochs followed by a cosine decay schedule without restarts \cite{loshchilov2016sgdr}. All the implementations were done in Python using the PyTorch\footnote{\url{https://pytorch.org/}} framework.\\
 
\subsubsection*{\textbf{Evaluation Datasets}}
We ran experiments on 8 large IQA databases spanning both synthetic and authentic distortions. 
\begin{itemize}
    \item Authentic Distortions
    \begin{itemize}
        \item KonIQ \cite{hosu2020koniq} : contains 10k images sampled from the public media database YFCC100M \cite{thomee2016yfcc100m}.
        \item CLIVE \cite{ghadiyaram2015massive} : contains 1162 images captured from many diverse mobile devices.
        \item FLIVE \cite{ying2019patches} : contains 40k real-world images and 120k patches along with respective quality scores. We only used images (and their corresponding scores) for analysis, and did not include patch information. 
        \item SPAQ \cite{fang2020perceptual} : contains 11k images captured using 66 smartphones. We only used images and their corresponding scores, and did not utilize the additional tag information available. Similar to \cite{fang2020perceptual}, we resized the images before evaluation such that the shorter side is 512.
    \end{itemize}
    \item Synthetic Distortions
    \begin{itemize}
        \item LIVE-IQA \cite{sheikh2006statistical} : contains 779 distorted images obtained from 29 pristine images using 5 synthetic distortion types.
        \item CSIQ-IQA \cite{larson2010most} : contains 866 distorted images obtained from 30 source contents with 6 types of distortions.
        \item TID2013 \cite{ponomarenko2015image} : contains 3000 distorted images obtained from 25 natural images with 24 distortion types, each having 5 levels of degradation.
        \item KADID \cite{lin2019kadid} : contains 10125 distorted images from 81 source contents spanning 25 different types of distortions.
    \end{itemize}
\end{itemize}

\subsubsection*{\textbf{Compared Methods}}
We compare the performance of CONTRIQUE against nine SOTA NR IQA models. The compared methods can be categorized into 3 categories : (a) Traditional/hand-crafted features - BRISQUE \cite{mittal2012no} and NIQE \cite{mittal2013making}. (b) Codebook-based features - CORNIA \cite{ye2012unsupervised} and HOSA \cite{xu2016blind}. Except NIQE, the rest use a support vector regressor (SVR) for quality prediction. (c) CNN based models - DB-CNN \cite{zhang2018blind}, PQR \cite{zeng2017probabilistic}, BIECON \cite{kim2016fully}, PaQ-2-PiQ \cite{ying2019patches} and HyperIQA \cite{su2020blindly}. For objective comparison of the above IQA models, we copied the numbers as reported by the respective authors or as available in the literature. For PaQ-2-PiQ, we consider the baseline model, since patch quality scores are not employed for training. We also included a Resnet-50 \cite{he2016deep} model pretrained on Imagenet \cite{russakovsky2015imagenet}, using a similar linear regression module as CONTRIQUE to predict quality. This comparison enabled us to compare the effect of supervised and unsupervised pretraining techniques. 

\subsubsection*{\textbf{Evaluation Protocol}}
Two commonly used evaluation metrics Spearman's rank order correlation coefficient (SROCC) and Pearson's linear correlation coefficient (PLCC) were employed to evaluate and compare the IQA models. Before computing PLCC, the quality predictions were passed through a four-parameter logistic non-linearity as described in \cite{VQEG2000}.

Each dataset was randomly divided into 70\%, 10\% and 20\% corresponding to training, validation and test sets, respectively. The validation set was used to determine the regularization coefficient of the regressor  using grid search. On datasets with synthetic distortions, the splits were implemented based on reference images, to ensure no overlap of contents. To avoid any bias towards the choice of training set, we repeated the train/test split operation 10 times and reported the median performance. On FLIVE, due to the large size of the dataset, we used a single split as reported by the authors in \cite{ying2019patches}.

\subsection{Correlation Against Human Judgments}
\label{sec:correlation}

\begin{table*}[t]
\caption{Performance comparison of CONTRIQUE against different NR models on IQA databases containing \textbf{synthetic} distortions. Models are categorized based on the type of feature extraction used. In each column, the first and second best models are boldfaced. Entries marked '-' denote that the results are not available.}
\label{table:synthetic_IQA}
    \centering
    \footnotesize
    \begin{tabular}{|c||c||cc|cc|cc|cc|}
        \hline
        \multirow{2}{*}{Method} & \multirow{2}{*}{Model Type} & \multicolumn{2}{|c|}{LIVE-IQA \cite{sheikh2006statistical}} & \multicolumn{2}{|c|}{CSIQ-IQA \cite{larson2010most}} & \multicolumn{2}{|c|}{TID2013 \cite{ponomarenko2015image}} & \multicolumn{2}{|c|}{KADID \cite{lin2019kadid}} \\
        \cline{3-10}
        ~ & ~ & SROCC$\uparrow$ & PLCC$\uparrow$ & SROCC$\uparrow$ & PLCC$\uparrow$ & SROCC$\uparrow$ & PLCC$\uparrow$ & SROCC$\uparrow$ & PLCC$\uparrow$\\ \hline \hline
        BRISQUE \cite{mittal2012no} & \multirow{2}{*}{\shortstack[c]{Traditional/Handcrafted \\ Features}} & 0.939 & 0.935 & 0.746 & 0.829 & 0.604 & 0.694 & 0.528 & 0.567\\
        NIQE \cite{mittal2013making} & ~ & 0.907 & 0.901 & 0.627 & 0.712 & 0.315 & 0.393 & 0.374 & 0.428\\ \hline
        CORNIA \cite{ye2012unsupervised} & \multirow{2}{*}{\shortstack[c]{Codebook-based \\ Features}} & 0.947 & 0.950 & 0.678 & 0.776 & 0.678 & 0.768 & 0.516 & 0.558 \\
        HOSA \cite{xu2016blind} & ~ & 0.946 & 0.950 & 0.741 & 0.823 & 0.735 & 0.815 & 0.618 & 0.653 \\ \hline
        DB-CNN \cite{zhang2018blind} & \multirow{4}{*}{\shortstack[c]{Supervised pretraining and \\ supervised fine-tuning}} & \textbf{0.968} & \textbf{0.971} & \textbf{0.946} & \textbf{0.959} & 0.816 & \textbf{0.865} & 0.851 & \textbf{0.856}\\
        PQR \cite{zeng2017probabilistic} & ~ & \textbf{0.965} & \textbf{0.971} & 0.872 & 0.901 & 0.740 & 0.798 & - & -\\
        BIECON \cite{kim2016fully} & ~ & 0.961 & 0.962 & 0.815 & 0.823 & 0.717 & 0.762 & - & - \\ 
        HyperIQA \cite{su2020blindly} & ~ & 0.962 & 0.966 & 0.923 & 0.942 & \textbf{0.840} & \textbf{0.858} & \textbf{0.852} & 0.845 \\\hline
        \multirow{2}{*}{Resnet-50 \cite{he2016deep}} & \multirow{2}{*}{\shortstack[c]{Supervised pretraining and \\ Linear Regression}} & \multirow{2}{*}{0.925} & \multirow{2}{*}{0.931} & \multirow{2}{*}{0.840} & \multirow{2}{*}{0.848} & \multirow{2}{*}{0.679} & \multirow{2}{*}{0.729} & \multirow{2}{*}{0.701} & \multirow{2}{*}{0.677} \\ 
        ~ & ~ & ~ & ~ & ~ & ~ & ~ & ~ & ~ & ~\\ \hline
        \multirow{2}{*}{CONTRIQUE} & \multirow{2}{*}{\shortstack[c]{Unsupervised pretraining and \\ Linear Regression}} & \multirow{2}{*}{0.960} & \multirow{2}{*}{0.961} & \multirow{2}{*}{\textbf{0.942}} & \multirow{2}{*}{\textbf{0.955}} & \multirow{2}{*}{\textbf{0.843}} & \multirow{2}{*}{0.857} & \multirow{2}{*}{\textbf{0.934}} & \multirow{2}{*}{\textbf{0.937}} \\
        ~ & ~ & ~ & ~ & ~ & ~ & ~ & ~ & ~ & ~ \\
        \hline
    \end{tabular}
\end{table*}

We compared the performance of CONTRIQUE against other models on IQA datasets containing authentic distortions in Table \ref{table:authentic_IQA}. It may be observed from the table that CONTRIQUE achieves competitive performance when compared to other SOTA models. In the table, we categorized the models based on the type of feature extraction techniques. Notably CONTRIQUE achieves performance comparable to CNN based fine-tuned models even without fine-tuning, highlighting the effectiveness of our proposed self-supervision methodology. Furthermore, it outperformed Resnet-50 features, reinforcing the efficiency of the auxiliary task employed in CONTRIQUE.

In Table \ref{table:synthetic_IQA} model performances are compared on datasets with synthetic distortions. Here as well, CONTRIQUE achieved superior performance among the compared models, indicating a better generalizability of learned representations across both synthetic and authentic distortions.

\subsection{Cross Dataset Evaluation}
\begin{table}[t]
\caption{Cross database SROCC comparison of IQA models. In each row top performing model is highlighted.}
\label{table:cross_data}
    \centering
    \scriptsize
        \begin{tabular}{|c|c||c|c|c|c|}
        \hline
        Training & Testing & DB-CNN & PQR & HyperIQA & CONTRIQUE \\ \hline
        CLIVE & KonIQ & 0.754 & 0.757 & \textbf{0.772} & 0.676 \\ 
        KonIQ & CLIVE & 0.755 & 0.770 & \textbf{0.785} & 0.731 \\ 
        LIVE-IQA & CSIQ-IQA & 0.758 & 0.719 & 0.744 & \textbf{0.823} \\ 
        CSIQ-IQA & LIVE-IQA & 0.877 & 0.922 & \textbf{0.926} & \textbf{0.925} \\
        \hline
    \end{tabular}
\end{table}

We conducted cross dataset evaluations whereby training and testing was performed on different datasets to analyze the dependence of training data, yielding the results reported in Table \ref{table:cross_data}. For simplicity we only include 4 datasets for comparison, two each from synthetic and realistic distortion sets. It may be inferred from the table that CONTRIQUE attains performance comparable to other IQA models across both synthetic and authentic distortions. Note that for CONTRIQUE, even for cross-dataset evaluations, only the weights of the linear regressor are modified depending on the training data, while the weights of the encoder backbone were kept intact.

\subsection{Visual Comparison of Representations}

\begin{figure}[t]
    \centering
    \subfloat[CONTRIQUE]{\includegraphics[width = 0.48\linewidth]{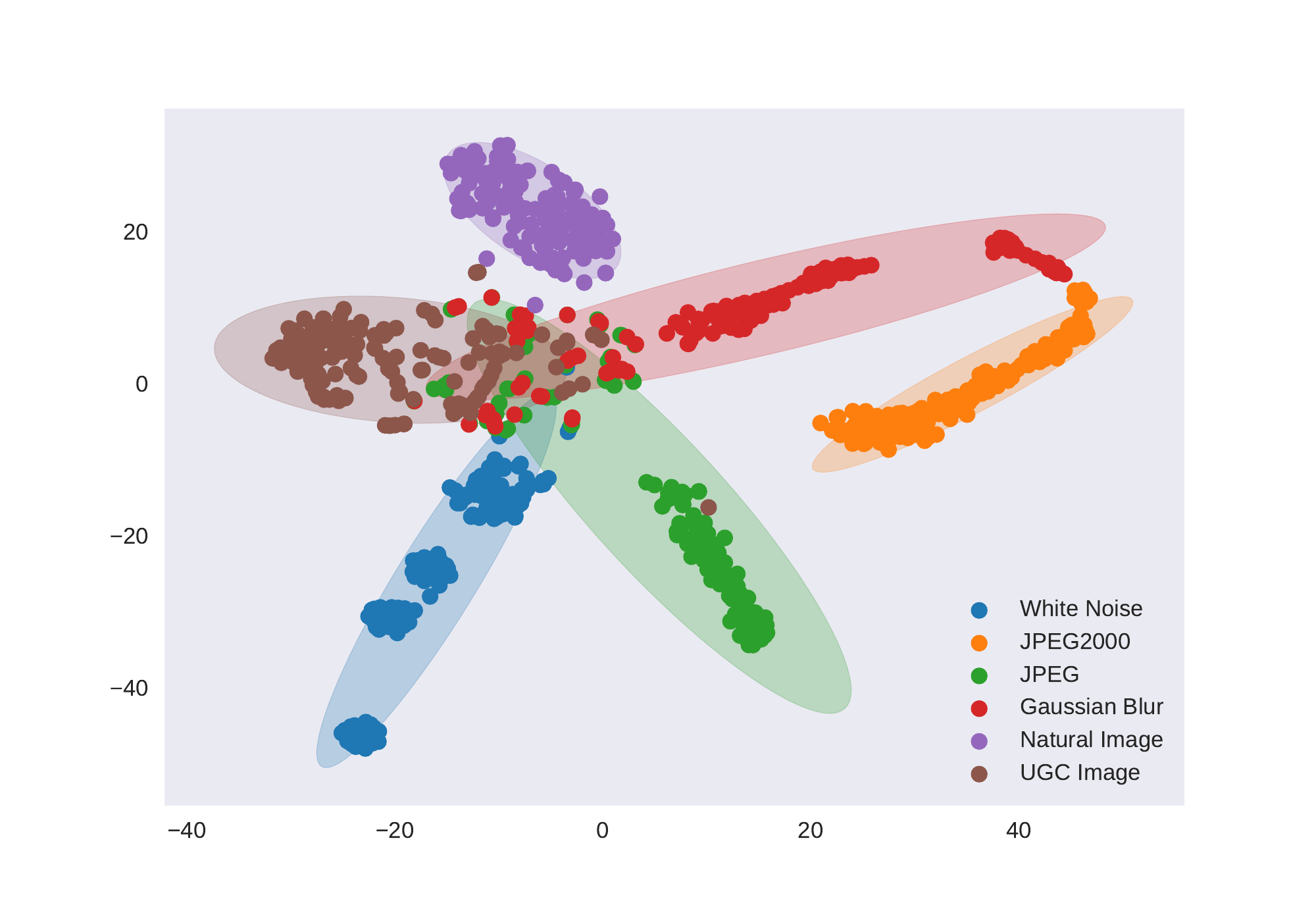}} \quad
    \subfloat[Resnet-50]{\includegraphics[width=0.48\linewidth]{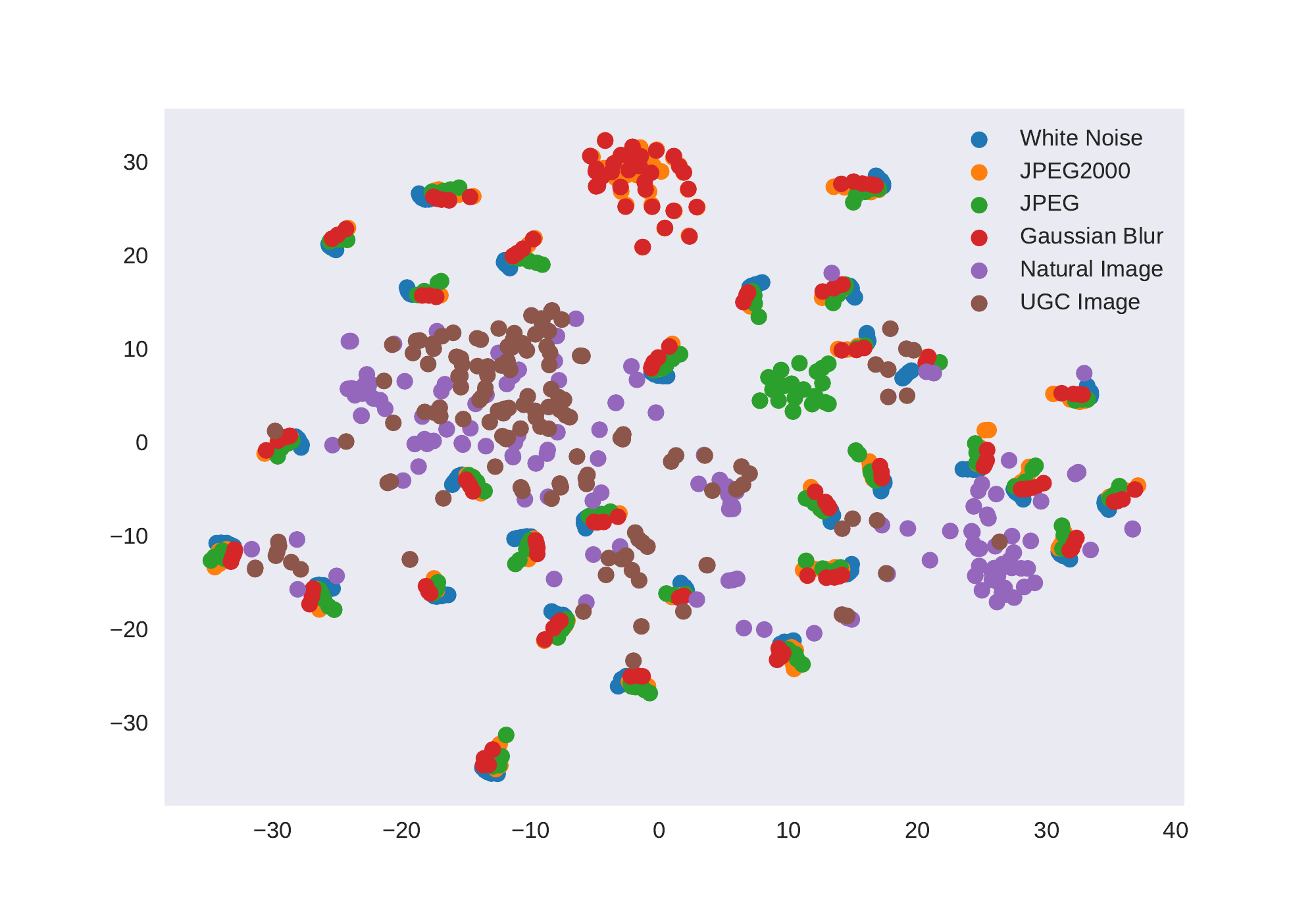}} \quad
    \subfloat[White Noise]{\includegraphics[width = 0.48\linewidth]{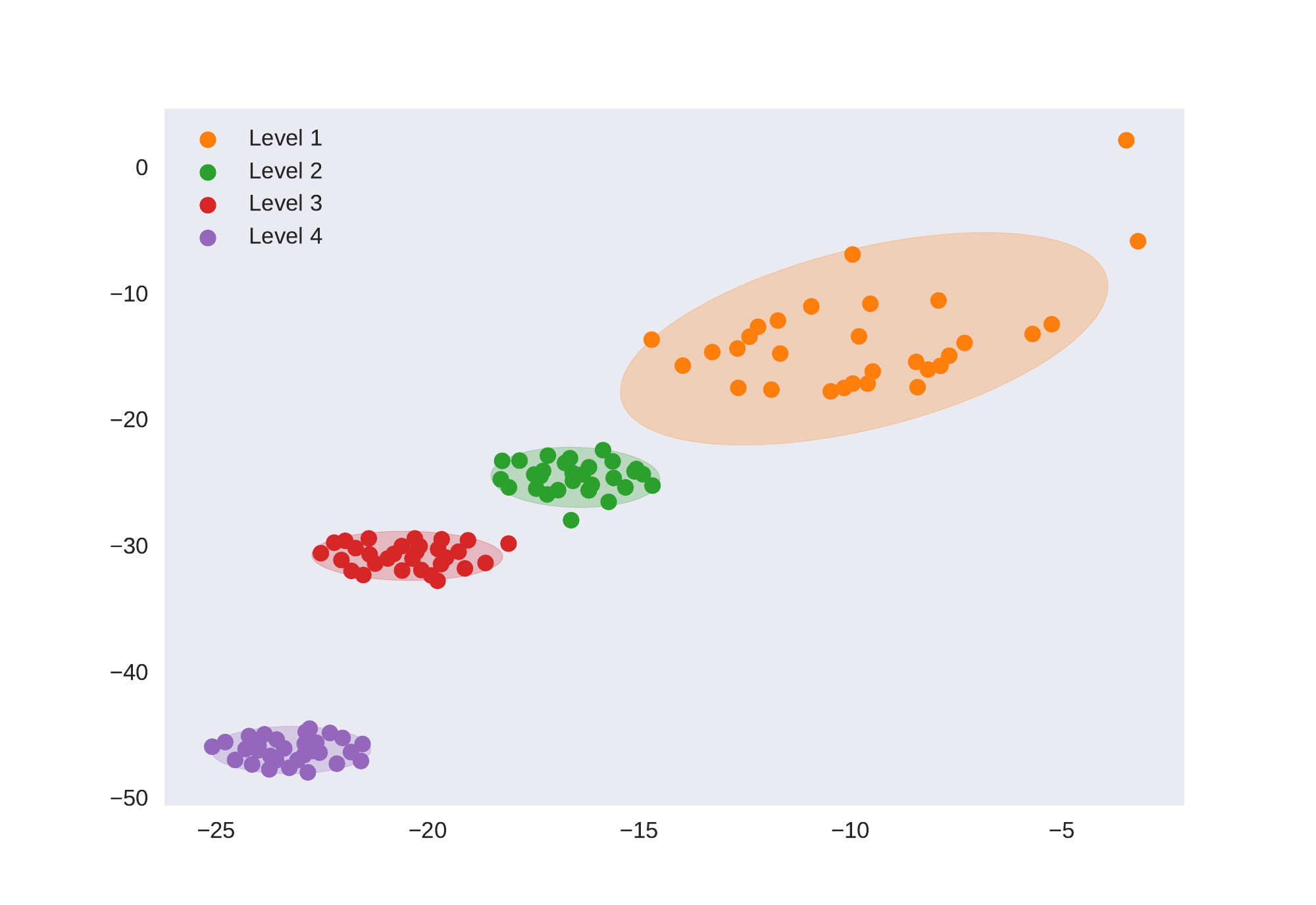}} \quad
    \subfloat[JPEG Compression]{\includegraphics[width = 0.48\linewidth]{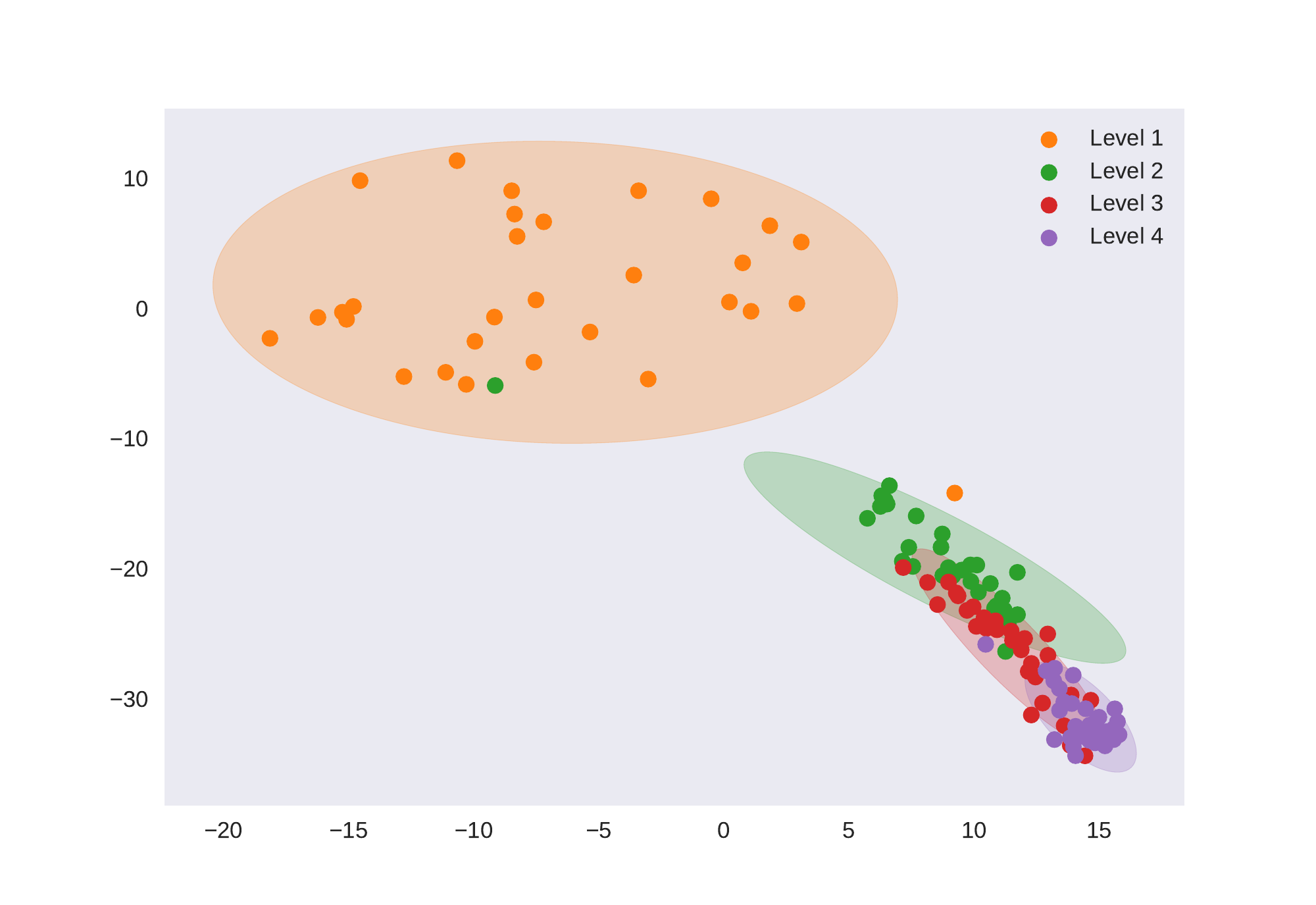}}
    \caption{Visualization of learned representations. (c) and (d) are zoomed versions of the white noise and JPEG compression clusters shown in (a).}
    \label{fig:comprison_representations}
\end{figure}

The learned representations for CONTRQUE are visualized in Fig. \ref{fig:comprison_representations} using t-sne \cite{van2008visualizing}. In the figure, for plotting purposes we used 4 commonly observed synthetic distortions: white noise, Gaussian blur, JPEG, and JPEG200, along with natural and UGC images. Each set contains 150 images, with synthetic distortions taken from the CSIQ-IQA dataset, while natural and UGC images sampled from the KADIS and KonIQ datasets, respectively. For comparison, we also include features from a Resnet-50 (Imagenet pretrained) in Fig. \ref{fig:comprison_representations}. Since the auxiliary task was to learn distortion discriminable embeddings, the learned CONTRIQUE features can be easily clustered depending on the type of distortions, as shown in Fig. \ref{fig:comprison_representations}. However, the same is not true of Resnet-50 features, as they appear to be scattered across the space and did not form separable clusters. Fig. \ref{fig:comprison_representations} also illustrates the degradation level separability of CONTRIQUE features for the white noise and JPEG compression distortions.

\begin{table*}[t]
\centering
    \caption{Full Reference performance comparison across 4 IQA databases. In each column, the first and second best models are boldfaced. Entries marked '-' denote that the results are not available.}
    \label{table:FR_methods_comparison}
    \begin{tabular}{|c||c|c|c|c|c|c|c|c|}
        \hline
        \multirow{2}{*}{Method}& \multicolumn{2}{|c|}{LIVE-IQA\cite{sheikh2006statistical}} & \multicolumn{2}{|c|}{CSIQ-IQA \cite{larson2010most}} & \multicolumn{2}{|c|}{TID2013 \cite{ponomarenko2015image}} & \multicolumn{2}{|c|}{KADID \cite{lin2019kadid}} \\
        \cline{2-9}
        ~ & SROCC$\uparrow$ & PLCC$\uparrow$ & SROCC$\uparrow$ & PLCC$\uparrow$ & SROCC$\uparrow$ & PLCC$\uparrow$ & SROCC$\uparrow$ & PLCC$\uparrow$ \\ \hline \hline
        PSNR & 0.881 & 0.868 & 0.820 & 0.824 & 0.643 & 0.675 & 0.677 & 0.680 \\ 
        SSIM \cite{wang2004image} & 0.921 & 0.911 & 0.854 & 0.835 & 0.642 & 0.698 & 0.641 & 0.633 \\  
        FSIM \cite{zhang2011fsim} & 0.964 & 0.954 & 0.934 & 0.919 & 0.852 & 0.875 & 0.854 & 0.850 \\
        VSI \cite{zhang2014vsi} & 0.951 & 0.940 & 0.944 & 0.929 & 0.902 & 0.903 & \textbf{0.880} & \textbf{0.878} \\ \hline
        PieAPP \cite{prashnani2018pieapp} & 0.915 & 0.905 & 0.900 & 0.881 & 0.877 & 0.850 & 0.869 & 0.869 \\ 
        LPIPS \cite{zhang2018unreasonable} & 0.932 & 0.936 & 0.884 & 0.906 & 0.673 & 0.756 & 0.721 & 0.713 \\
        DISTS \cite{ding2020image} & 0.953 & 0.954 & 0.942 & 0.942 & 0.853 & 0.873 & - & - \\ 
        DRF-IQA \cite{kim2020dynamic} & \textbf{0.983} & \textbf{0.983} & \textbf{0.964} & \textbf{0.960} & \textbf{0.944} & \textbf{0.942} & - & - \\ \hline
        CONTRIQUE-FR & \textbf{0.966} & \textbf{0.966} & \textbf{0.956} & \textbf{0.964} & \textbf{0.909} & \textbf{0.915} & \textbf{0.946} & \textbf{0.947} \\  
        \hline
    \end{tabular}
\end{table*}

\subsection{Significance of Training Data}
\begin{table}[t]
    \caption{SROCC performance comparison of different trainings of CONTRIQUE. \textit{syn} and \textit{UGC} denote models trained with data containing only synthetic and authentic distortions respectively. In each column, the top performing model is boldfaced.}
    \label{table:training_data}
    \centering
    \footnotesize
        \begin{tabular}{|c||c|c|c|c|}
        \hline
        Model & KonIQ & CLIVE & LIVE-IQA & CSIQ-IQA \\ \hline \hline
        CONTRIQUE-syn & 0.854 & 0.756 & \textbf{0.965} & \textbf{0.950} \\ 
        CONTRIQUE-UGC & \textbf{0.900} & \textbf{0.843} & 0.918 & 0.802 \\ \hline
        CONTRIQUE & 0.894 & \textbf{0.846} & 0.960 & 0.942 \\
        \hline
    \end{tabular}
\end{table}

During training of CONTRIQUE, we employed a mixed dataset containing both synthetic and realistic distortions. We conducted an ablation study whereby the effects of synthetic and authentic distortions were analyzed in isolation. In this experiment, CONTRIQUE was trained with data containing either only synthetic or authentic artifacts, with the performance numbers reported in Table \ref{table:training_data}. From the Table, we can infer that training with only synthetic distortions boosts performance on synthetic IQA datasets, while the same holds true for authentic IQA datasets when trained on UGC data. Employing mixed data achieves better generalization, with negligible loss in performance as compared to the individual trainings.

\subsection{Importance of Different Color Spaces}
\label{sec:results_colorspace}
In the CONTRIQUE, different color spaces were employed in order to extract complementary quality information. In this experiment we investigate the significance of each color space by training CONTRIQUE on each of them individually. The results are reported in Table \ref{table:training_colorspace}, and it can be observed that combined training yields superior correlations than for any of the individual color spaces highlighting their complementary nature. Note that during evaluation, the images were converted to the respective color spaces on which they were trained.

\begin{table}[t]
    \caption{SROCC performance variation of CONTRIQUE for the different color spaces used during training. In each column the top performing model is boldfaced.}
    \label{table:training_colorspace}
    \centering
    \footnotesize
        \begin{tabular}{|c||c|c|c|c|}
        \hline
        Training & \multirow{2}{*}{KonIQ} & \multirow{2}{*}{CLIVE} & \multirow{2}{*}{LIVE-IQA} & \multirow{2}{*}{CSIQ-IQA} \\
        Color space & ~ & ~ & ~ & ~ \\ \hline \hline
        Grayscale & 0.837 & 0.758 & 0.948 & 0.861 \\
        RGB & 0.834 & 0.757 & 0.948 & 0.888 \\
        LAB & 0.737 & 0.600 & 0.819 & 0.75 \\
        HSV & 0.766 & 0.650 & 0.909 & 0.823 \\
        MS & 0.870 & 0.800 & \textbf{0.960} & 0.903 \\ \hline
        All & \textbf{0.894} & \textbf{0.846} & \textbf{0.960} & \textbf{0.942} \\
        \hline
    \end{tabular}
\end{table}

Another interesting observation we make is the invariance property of the learned CONTRIQUE representations to the different color spaces. In other words, during evaluation, using any color space yielded approximately similar embeddings. This behavior is illustrated in Table \ref{table:testing_colorspace}, where during evaluation the images were converted to multiple color spaces. From the Table it can be inferred that the performances of CONTRIQUE remained approximately same across color spaces. This property is a consequence of using multiple color spaces during training. Furthermore, this property eliminates the need of changing color spaces during evaluation without significantly sacrificing performance.

\section{CONTRIQUE Full-Reference Model}
\begin{table}[t]
    \caption{SROCC performance variation of CONTRIQUE when evaluated on different color spaces.}
    \label{table:testing_colorspace}
    \centering
    \footnotesize
        \begin{tabular}{|c||c|c|c|c|}
        \hline
        Testing & \multirow{2}{*}{KonIQ} & \multirow{2}{*}{CLIVE} & \multirow{2}{*}{LIVE-IQA} & \multirow{2}{*}{CSIQ-IQA} \\
        Color space & ~ & ~ & ~ & ~ \\ \hline \hline
        Grayscale & 0.891 & 0.846 & 0.960 & 0.911 \\
        RGB & 0.894 & 0.846 & 0.960 & 0.942 \\
        LAB & 0.886 & 0.838 & 0.953 & 0.921 \\
        HSV & 0.889 & 0.843 & 0.960 & 0.941 \\
        MS & 0.880 & 0.821 & 0.955 & 0.881 \\
        \hline
    \end{tabular}
\end{table}
CONTRIQUE framework offers the flexibility to employ the learned representations on other IQA related tasks. We also propose a simple extension to employ CONTRIQUE representations in a Full (FR) IQA setting, where we have access to both pristine high quality reference images as well as their corresponding distorted versions. To incorporate reference information into the regressor, equation (\ref{eqn:regressor}) is modified as

\begin{align}
\begin{aligned}
    y &= W |h_{ref} - h_{dist}|, \\ 
    W^* &= \argmin_{W} \sum_{i=1} ^N (GT_i - y_i)^2 + \lambda \sum_{j=1} ^M W_j^2,
\end{aligned}
\end{align}
where absolute difference between the features of reference and distorted images are used to predicting quality. We denote this modified model as CONTRIQUE-FR. Note that we do not perform any additional training or fine-tuning of the encoder network for the FR-IQA task. The same trained encoder obtained from CONTRIQUE was used with only the regressor modified to include reference information.

The performance of CONTRIQUE-FR is compared in Table \ref{table:FR_methods_comparison}. We followed a similar evaluation protocol of dividing datasets into 70\%/10\%/20\% as train/validation/test sets, respectively based on content, and report the median correlation values over 10 different train/test splits. Since authentic IQA datasets do not contain reference images, we only report performances on the synthetic IQA datasets. For comparison, we include eight SOTA FR-IQA models : (a) Traditional models - PSNR, SSIM \cite{wang2004image}, FSIM \cite{zhang2011fsim} and VSI \cite{zhang2014vsi}. (b) Deep learning based models - PieAPP \cite{prashnani2018pieapp}, LPIPS \cite{zhang2018unreasonable}, DISTS \cite{ding2020image} and DRF-IQA \cite{kim2020dynamic}. From the Table it can be observed that CONTRIQUE-FR achieves performance comparable to SOTA FR-IQA models, highlighting the flexibility as well as generalizability of the CONTRIQUE training framework. Additionally, comparing the CONTRIQUE correlation values in Table \ref{table:synthetic_IQA} and \ref{table:FR_methods_comparison} shows the performance gains due to the knowledge of the high quality reference images.

\section{Conclusion}
\label{sec:conclusion}
We introduced an unsupervised training framework that learns effective image quality representations. Distinguishing characteristics of the proposed design include learning from unlabeled data, and employing distortion type and degree discrimination as an auxiliary task. We conducted holistic evaluations of our proposed model across multiple IQA databases, and found that CONTRIQUE achieves competitive performance against other, supervised IQA models. The proposed framework is simple, achieves superior performance with no additional fine-tuning, and generalizes well across synthetic and realistic distortions. We conducted ablation experiments to understand the significance of different color spaces, and found surprisingly complementary quality prediction power among them. We also analyzed the importance of the distortion types present in the training data, and deduced that using a combination of synthetic and authentic artifacts helps achieve better generalization. We also proposed CONTRIQUE-FR, an extension of CONTRIQUE to FR IQA problem, which required no additional training of the CNN backbone. CONTRIQUE-FR also achieved comparable performance against SOTA FR-IQA models. A software release of CONTRIQUE and CONTRiQUE-FR has been made available online\footnote{\url{https://github.com/pavancm/CONTRIQUE}}.

\section{Acknowledgment}
The authors would like to thank YouTube for supporting this research, and the Texas Advanced Computing Center (TACC) for providing computational resources that contributed to this work. This work was also supported by grant number 2019844 for the National Science Foundation AI Institute for Foundations of Machine Learning (IFML).

\bibliographystyle{IEEEtran}
\bibliography{references}

\end{document}